\documentclass[11pt,a4paper]{article}
\usepackage[hyperref]{naaclhlt2019}
\usepackage{times}
\usepackage{latexsym}
\usepackage{colortbl}
\usepackage{amssymb}
\usepackage{amsmath}

\usepackage[pdftex]{graphicx}

\DeclareMathOperator*{\argmax}{arg\,max}

\usepackage[tight,footnotesize]{subfigure}

\usepackage{url}

\aclfinalcopy 
\def\aclpaperid{619} 

\title{Neural Grammatical Error Correction with Finite State Transducers}

\author{Felix Stahlberg$^\dagger$ \and Christopher Bryant$^\ddagger$ \and Bill Byrne$^\dagger$ \\
  $^\dagger$Department of Engineering \\
  $^\ddagger$ Department of Computer Science and Technology \\
  University of Cambridge \\
  United Kingdom \\
  {\tt \{fs439, cjb255, wjb31\}@cam.ac.uk}\\
}

\date{}

\begin{document}
\maketitle
\begin{abstract}
Grammatical error correction (GEC) is one of the areas in natural language processing in which purely neural models have not yet superseded more traditional symbolic models. Hybrid systems combining phrase-based statistical machine translation (SMT) and neural sequence models are currently among the most effective approaches to GEC. However, both SMT and neural sequence-to-sequence models require large amounts of annotated data. Language model based GEC (LM-GEC) is a promising alternative which does not rely on annotated training data. We show how to improve LM-GEC by applying modelling techniques based on finite state transducers. We report further gains by rescoring with neural language models. We show that our methods developed for LM-GEC can also be used with SMT systems if annotated training data is available. Our best system outperforms the best published result on the CoNLL-2014 test set, and achieves far better relative improvements over the SMT baselines than previous hybrid systems.
\end{abstract}

\section{Introduction}

Grammatical error correction (GEC) is the task of automatically correcting all types of errors in text; e.g. [\textit{In a such situaction} $\rightarrow$ \textit{In such a situation}]. Using neural models for GEC is becoming increasingly popular~\citep{neural-first,neural-cam,neural-nested-att,neural-rl,neural-harvard,neural-conv,neural-fluency,ge-human}, possibly combined with phrase-based SMT~\citep{ntm-in-smt,char-smt,marcin2018}. A potential challenge for purely neural GEC models is their vast output space since they assign non-zero probability mass to any sequence. GEC is -- compared to machine translation -- a highly constrained problem as corrections tend to be very local, and lexical choices are usually limited. Finite state transducers (FSTs) are an efficient way to represent large structured search spaces. In this paper, we propose to construct a hypothesis space using standard FST operations like composition, and then constrain the output of a neural GEC system to that space. We study two different scenarios: In the first scenario, we do not have access to annotated training data, and only use a small development set for tuning. In this scenario, we construct the hypothesis space using word-level context-independent confusion sets~\citep{chris-lm} based on spell checkers and morphology databases, and rescore it with count-based and neural language models (NLMs). In the second scenario, we assume to have enough training data available to train SMT and neural machine translation (NMT) systems. In this case, we make additional use of the SMT lattice and rescore with an NLM-NMT ensemble. Our contributions are:

\begin{itemize}
    \item We present an FST-based adaptation of the work of \citet{chris-lm} which allows exact inference, and does not require annotated training data. We report large gains from rescoring with a neural language model.
    \item Our technique beats the best published result with comparable amounts of training data on the CoNLL-2014~\citep{conll2014} test set when applied to SMT lattices. Our combination strategy yields larger gains over the SMT baselines than simpler rescoring or pipelining used in prior work on hybrid systems~\citep{marcin2018}.
\end{itemize}

\section{Constructing the Hypothesis Space}

\paragraph{Constructing the set of hypotheses}
\label{sec:hypo-space-construction}

\begin{figure*}[!t]
\centering
\small
\subfigure[The input lattice $I$ without SMT (no annotated training data).]{\label{fig:build_hypo_i_nosmt}\includegraphics[width=0.70\linewidth]{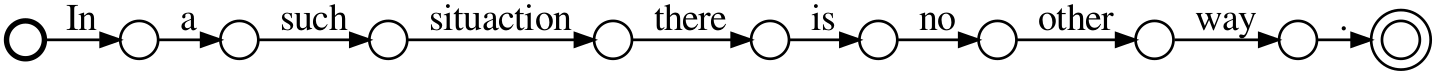}}
\subfigure[The base lattice $B$ without SMT.]{\label{fig:build_hypo_b_nosmt}\includegraphics[width=1\linewidth]{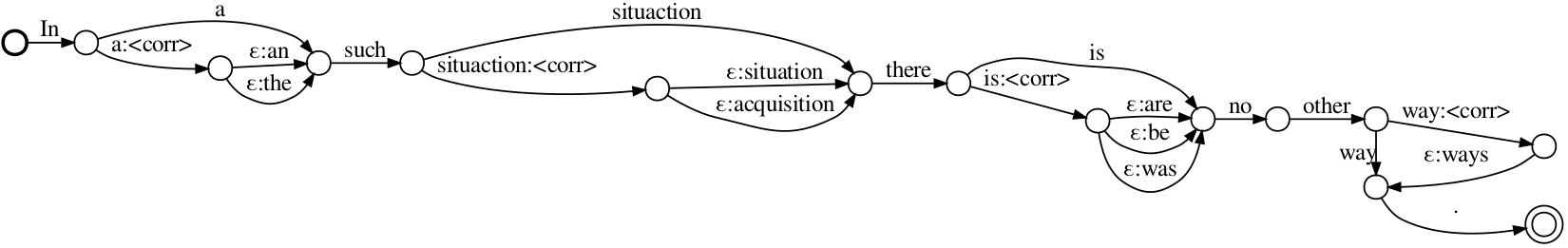}}
\subfigure[The input lattice $I$ with SMT.]{\label{fig:build_hypo_i_smt}\includegraphics[width=1\linewidth]{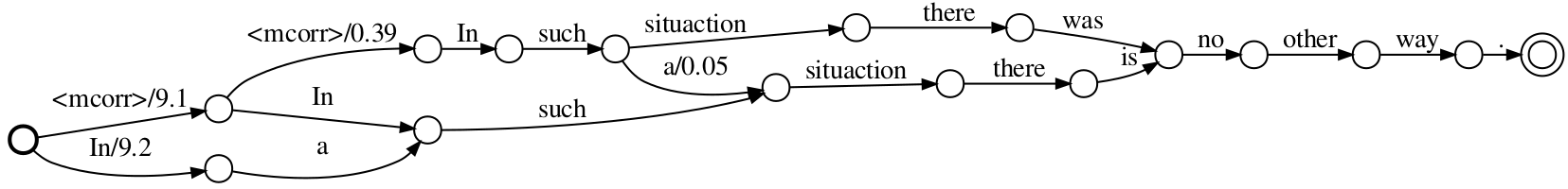}}
\subfigure[The base lattice $B$ with SMT.]{\label{fig:build_hypo_b}\includegraphics[width=1\linewidth]{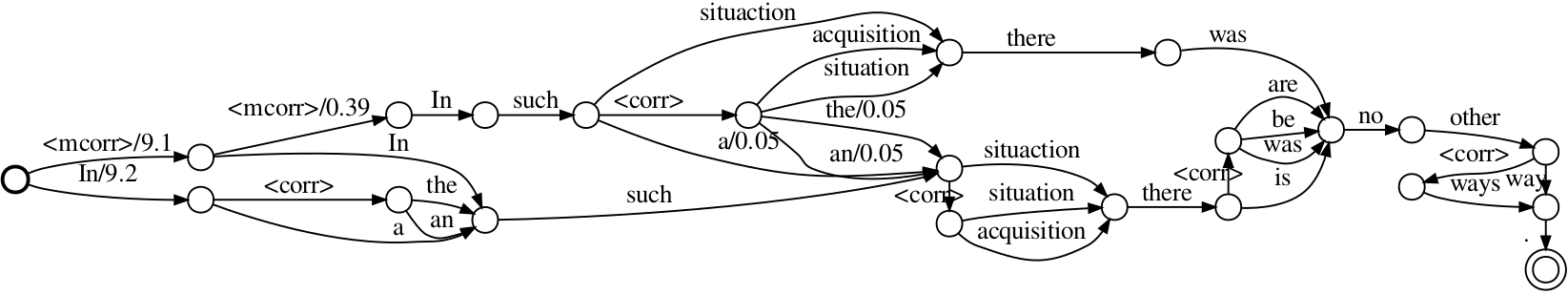}}
\caption{Building the hypothesis space for the input sentence ``In a such situaction there is no other way .''.}
\label{fig:build_hypo}
\end{figure*}

\begin{figure}[!b]
\centering
\small
\includegraphics[width=0.57\linewidth]{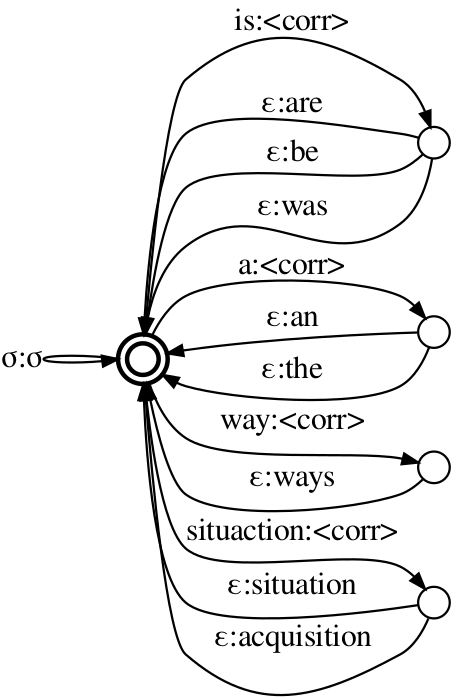}
\caption{The edit transducer $E$. The $\sigma$-label can match any input symbol.}
\label{fig:edit_fst}
\end{figure}

\begin{figure}[!b]
\centering
\small
\includegraphics[width=0.75\linewidth]{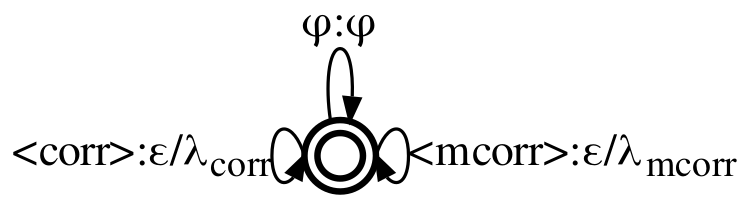}
\caption{The penalization transducer $P$. The $\phi$-label can match any input except $\mathtt{\textless corr\textgreater}$ and $\mathtt{\textless mcorr\textgreater}$.}
\label{fig:pen_fst}
\end{figure}

The core idea of our approach is to first construct a (weighted) hypothesis space $H$ which is large enough to be likely to contain good corrections, but constrained enough to embrace the highly structured nature of GEC. Then, we use $H$ to constrain a neural beam decoder. We make extensive use of the FST operations available in OpenFST~\citep{openfst} like composition (denoted with the $\circ$-operator) and projection (denoted with $\Pi_\text{input}(\cdot)$ and $\Pi_\text{output}(\cdot)$) to build $H$. The process starts with an input lattice $I$. In our experiments without annotated training data, $I$ is an FST which simply maps the input sentence to itself as shown in Fig.~\ref{fig:build_hypo_i_nosmt}. If we do have access to enough annotated data, we train an SMT system on it and derive $I$ from the SMT $n$-best list.\footnote{In the rare cases in which the $n$-best list did not contain the source sentence $\mathbf{x}$ we added it in a postprocessing step.} For each hypothesis $\mathbf{y}$ we compute the Levenshtein distance $\text{lev}(\mathbf{x},\mathbf{y})$ to the source sentence $\mathbf{x}$. We construct a string $\mathbf{z}$ by prepending $\text{lev}(\mathbf{x},\mathbf{y})$ many $\mathtt{\textless mcorr\textgreater}$ tokens to $\mathbf{y}$, and construct $I$ such that:
\begin{align}
\mathbf{z} = {(\mathtt{\textless mcorr\textgreater})}^{\text{lev}(\mathbf{x},\mathbf{y})}\cdot \mathbf{y} \\
[[I]](\mathbf{z}) = -\lambda_\text{SMT} \text{SMT}(\mathbf{y}|\mathbf{x}).
\end{align}
We adapt the notation of \citet{mohri-edit} and denote the cost $I$ assigns to mapping a string $\mathbf{z}$ to itself as $[[I]](\mathbf{z})$, and set $[[I]](\mathbf{z})=\infty$ if $I$ does not accept $\mathbf{z}$. $\text{SMT}(\mathbf{y}|\mathbf{x})$ is the SMT score. In other words, $I$ represents the weighted SMT $n$-best list after adding $\text{lev}(\mathbf{x},\mathbf{y})$ many $\mathtt{\textless mcorr\textgreater}$ tokens to each hypothesis as illustrated in Fig.~\ref{fig:build_hypo_i_smt}. We scale SMT scores by a factor $\lambda_\text{SMT}$ for tuning.

\citet{chris-lm} addressed substitution errors such as non-words, morphology-, article-, and preposition-errors by creating confusion sets $C(x_i)$ that contain possible (context-independent) 1:1 corrections for each input word $x_i$. Specifically, they relied on CyHunspell for spell checking \citep{cyhunspell}, the AGID morphology database for morphology errors \citep{agid}, and manually defined confusion sets for determiner and preposition errors, hence avoiding the need for annotated training data. We use the same confusion sets as \citet{chris-lm} to augment our hypothesis space via the edit flower transducer $E$ shown in Fig.~\ref{fig:edit_fst}. $E$ can map any sequence to itself via its $\sigma$-self-loop. Additionally, it allows the mapping $x_i \rightarrow \mathtt{\textless corr\textgreater} \cdot y$ for each $y\in C(x_i)$. For example, for the misspelled word $x_i=\text{`situaction'}$ and the confusion set $C(\text{`situaction'}) = \{\text{`situation'},\text{`acquisition'}\}$, $E$ allows mapping `situaction' to `$\mathtt{\textless corr\textgreater}$ situation' and `$\mathtt{\textless corr\textgreater}$ acquisition', and to itself via the $\sigma$-self-loop. The additional $\mathtt{\textless corr\textgreater}$ token will help us to keep track of the edits. We obtain our {\em base lattice} $B$ which defines the set of possible hypotheses by composition and projection:
\begin{equation}
B:= \Pi_\text{output}(I\circ E).
\end{equation}
Fig.~\ref{fig:build_hypo_b} shows $B$ for our running example.

\paragraph{Scoring the hypothesis space}

We apply multiple scoring strategies to the hypotheses in $B$. First, we penalize $\mathtt{\textless mcorr\textgreater}$ and $\mathtt{\textless corr\textgreater}$ tokens with two further parameters, $\lambda_\text{mcorr}$ and $\lambda_\text{corr}$, by composing $B$ with the penalization transducer $P$ shown in Fig.~\ref{fig:pen_fst}.\footnote{Rather than using $\mathtt{\textless mcorr\textgreater}$ and $\mathtt{\textless corr\textgreater}$ tokens and the transducer $P$ we could directly incorporate the costs in the transducers $I$ and $E$, respectively. We chose to use explicit correction tokens for clarity.} The $\lambda_\text{mcorr}$ and $\lambda_\text{corr}$ parameters control the trade-off between the number and quality of the proposed corrections since high values bias towards fewer corrections.

To incorporate word-level language model scores we train a 5-gram count-based LM with KenLM~\citep{kenlm} on the One Billion Word Benchmark dataset~\citep{1b}, and convert it to an FST $L$ using the OpenGrm NGram Library~\citep{opengrm}. For tuning purposes we scale weights in $L$ with $\lambda_\text{KenLM}$:
\begin{equation}
[[L]](\mathbf{y}) = -\lambda_\text{KenLM} \log P_\text{KenLM}(\mathbf{y}).
\end{equation}
Our combined word-level scores can be expressed with the following transducer:
\begin{equation}
    H_\text{word}= B \circ P \circ L.
\end{equation}

Since we operate in the tropical semiring, path scores in $H_\text{word}$ are linear combinations of correction penalties, LM scores, and, if applicable, SMT scores, weighted with the $\lambda$-parameters. Note that exact inference in $H_\text{word}$ is possible using FST shortest path search. This is an improvement over the work of~\citet{chris-lm} who selected correction options greedily. Our ultimate goal, however, is to rescore $H_\text{word}$ with neural models such as an NLM and -- if annotated training data is available -- an NMT model. Since our neural models use subword units~\citep[BPEs]{bpe}, we compose $H_\text{word}$ with a transducer $T$ which maps word sequences to BPE sequences. Our final transducer $H_\text{BPE}$ which we use to constrain the neural beam decoder can be written as:
\begin{equation}
\begin{split}
H_\text{BPE} & = \Pi_\text{output}(H_\text{word} \circ T) \\
 & = \Pi_\text{output}(I\circ E \circ P \circ L \circ T).
\end{split}
\end{equation}
To help downstream beam decoding we apply $\epsilon$-removal, determinization, minimization, and weight pushing~\citep{mohri-lang,mohri-push} to $H_\text{BPE}$. We search for the best hypothesis $\mathbf{y}_\text{BPE}^*$ with beam search using a combined score of word-level symbolic models (represented by $H_\text{BPE}$) and subword unit based neural models:
\begin{equation}
\label{eq:y-star}
\begin{split}
\mathbf{y}_\text{BPE}^* = \argmax_{\mathbf{y}_\text{BPE}} \Big( &- [[H_\text{BPE}]](\mathbf{y}_\text{BPE}) \\
 &+ \lambda_\text{NLM}\log P_\text{NLM}(\mathbf{y}_\text{BPE}) \\
 &+ \lambda_\text{NMT}\log P_\text{NMT}(\mathbf{y}_\text{BPE}|\mathbf{x}_\text{BPE}) \Big)
\end{split}
\end{equation}

\begin{table*}
\centering
\small
\begin{tabular}{@{\hspace{0em}}r@{\hspace{0.2em}}|c|c|c||cccc|cccc|}
\cline{2-12}
 & {\bf Uses} & {\bf 5-gram} & {\bf NLM} & \multicolumn{4}{c|}{\bf CoNLL-2014} & \multicolumn{4}{c|}{\bf JFLEG Test} \\
 & {\bf $E$} & {\bf FST-LM} & {\bf (BPE)} & {\bf P} & {\bf R} & {\bf M2} & {\bf GLEU} & {\bf P} & {\bf R} & {\bf M2} & {\bf GLEU} \\\cline{2-12}
{\scriptsize 1} & \multicolumn{3}{|l||}{Best published (B\&B, \citeyear{chris-lm})}  & 40.56 & 20.81 & \cellcolor{gray!20}34.09 & 59.35 & 76.23 & 28.48 & \cellcolor{gray!20}57.08 & 48.75 \\\cline{2-12}
{\scriptsize 2} & \checkmark & \checkmark &  &  40.62 & 20.72 & \cellcolor{gray!20}34.08 & 64.03 & 81.08 & 28.69 & \cellcolor{gray!20}59.38 & 48.95 \\
{\scriptsize 3} & \checkmark & \checkmark &  \checkmark &  54.43 & 25.21 & \cellcolor{gray!20}44.19 & 66.75 & 79.88 &  32.99 & \cellcolor{gray!20}62.20 & 50.93 \\
{\scriptsize 4} & \checkmark & \checkmark &  \checkmark & 53.64 & 26.34 & 44.43 & \cellcolor{gray!20}66.89 & 70.24 & 38.94 & 60.51 & \cellcolor{gray!20}52.61 \\\cline{2-12}
\end{tabular}
\caption{Results without using annotated training data. Systems are tuned with respect to the metric highlighted in gray. Input lattices $I$ are derived from the source sentence as in Fig.~\ref{fig:build_hypo_i_nosmt}.}\label{tab:results-no-smt}
\end{table*}

\begin{table*}
\centering
\small
\begin{tabular}{@{\hspace{0em}}r@{\hspace{0.2em}}|c|c|c|c||cccc|cccc|}
\cline{2-13}
 & {\bf Uses} & {\bf 5-gram} & {\bf NMT} & {\bf NLM} & \multicolumn{4}{c|}{\bf CoNLL-2014} & \multicolumn{4}{c|}{\bf JFLEG Test} \\
 & {\bf $E$} & {\bf FST-LM} & {\bf (BPE)} & {\bf (BPE)} & {\bf P} & {\bf R} & {\bf M2} & {\bf GLEU} & {\bf P} & {\bf R} & {\bf M2} & {\bf GLEU} \\\cline{2-13}
 {\scriptsize 1} & \multicolumn{4}{|l||}{Best published (G\&J-D, \citeyear{marcin2018})} & 66.77 & 34.49 & \cellcolor{gray!20}56.25 & n/a & n/a & n/a & n/a & \cellcolor{gray!20}61.50 \\\cline{2-13}
{\scriptsize 2} & \multicolumn{4}{|l||}{Unconstrained single NMT} & 54.98 & 22.20 & 42.45 & 67.19 & 67.49 & 38.47 & 58.64 & 50.71 \\\cline{2-13}
{\scriptsize 3} & & &  &  & 60.95 & 26.21 & \cellcolor{gray!20}48.18 & 68.30 & 66.64 & 40.68 & \cellcolor{gray!20}59.09 & 50.86\\
{\scriptsize 4} & \checkmark & \checkmark &  &  & 57.58 & 32.39 & \cellcolor{gray!20}49.83 & 68.82 & 71.60 & 42.45  & \cellcolor{gray!20}62.95 & 53.20 \\
{\scriptsize 5} & &  &\checkmark  & \checkmark & 65.26 & 33.03 & \cellcolor{gray!20}54.61 & 69.92 & 76.35 & 40.55  & \cellcolor{gray!20}64.89 & 51.75 \\
{\scriptsize 6} & \checkmark &  &\checkmark  & \checkmark & 64.55 & 37.33 & \cellcolor{gray!20}56.33 & 70.30 & 78.85 & 47.72  & \cellcolor{gray!20}69.75 & 55.39 \\
{\scriptsize 7} & \checkmark &  &\checkmark (4x)  & \checkmark & 66.71 & 38.97 & \cellcolor{gray!20}58.40 & 70.60 & 82.15 & 47.82  & \cellcolor{gray!20}71.84 & 55.60 \\
{\scriptsize 8} & \checkmark &  &\checkmark (4x)  & \checkmark & 66.96 & 38.62 & 58.39 & \cellcolor{gray!20}70.60 & 74.19 & 56.41  & 69.79 & \cellcolor{gray!20}58.63 \\\cline{2-13}
\end{tabular}
\caption{Results with using annotated training data. Systems are tuned with respect to the metric highlighted in gray. Input lattices $I$ are derived from the Moses 1000-best list as in Fig.~\ref{fig:build_hypo_i_smt}. Row 3 is the SMT baseline.}\label{tab:results-smt}
\end{table*}

The final decoding pass can be seen as an ensemble of a neural LM and an NMT model which is constrained and scored at each time step by the set of possible tokens in $H_\text{BPE}$.


We have introduced three $\lambda$-parameters $\lambda_\text{corr}$, $\lambda_\text{KenLM}$, and $\lambda_\text{NLM}$, and three additional parameters $\lambda_\text{SMT}$, $\lambda_\text{mcorr}$, and $\lambda_\text{NMT}$ if we make use of annotated training data. We also use a word insertion penalty $\lambda_\text{wc}$ for our SMT-based experiments. We tune all these parameters on the development sets using Powell search~\citep{powell}.\footnote{Similarly to \citet{chris-lm}, even in our experiments without annotated {\em training} data, we do need a very small amount of annotated sentences for tuning.}

\section{Experiments}

\paragraph{Experimental setup}

In our experiments with annotated training data we use the SMT system of \citet{marcin2016}\footnote{\url{https://github.com/grammatical/baselines-emnlp2016}} to create 1000-best lists from which we derive the input lattices $I$. All our LMs are trained on the One Billion Word Benchmark dataset \citep{1b}. Our neural LM is a Transformer decoder architecture in the \texttt{transformer\_base} configuration trained with Tensor2Tensor~\citep{t2t}. Our NMT model is a Transformer model (\texttt{transformer\_base}) trained on the concatenation of the NUCLE corpus~\citep{nucle} and the Lang-8 Corpus of Learner English v1.0~\citep{lang8}. We only keep sentences with at least one correction (659K sentences in total). Both NMT and NLM models use byte pair encoding~\citep[BPE]{bpe} with 32K merge operations. We delay SGD updates by 2 on four physical GPUs as suggested by \citet{delayed-sgd}. We decode with beam size 12 using the SGNMT decoder~\citep{sgnmt}. We evaluate on CoNLL-2014 \citep{conll2014} and JFLEG-Test \citep{jfleg}, using CoNLL-2013 \citep{conll2013} and JFLEG-Dev as development sets. Our evaluation metrics are GLEU \citep{gleu} and M2 \citep{m2}. We generated M2 files using ERRANT \citep{errant} for JFLEG and Tab.~\ref{tab:results-no-smt} to be comparable to \citet{chris-lm}, but used the official M2 files in Tab.~\ref{tab:results-smt} to be comparable to \citet{marcin2018}.

\paragraph{Results}

Our LM-based GEC results without using annotated training data are summarized in Tab.~\ref{tab:results-no-smt}. Even when we use the same resources (same LM and same confusion sets) as \citet{chris-lm}, we see gains on JFLEG (rows 1 vs.\ 2), probably because we avoid search errors in our FST-based scheme. Adding an NLM  yields significant gains across the board. Tab.~\ref{tab:results-smt} shows that adding confusion sets to SMT lattices is effective even without neural models (rows 3 vs.\ 4). Rescoring with neural models also benefits from the confusion sets (rows 5 vs.\ 6). With our ensemble systems (rows 7 and 8) we are able to outperform prior work\footnote{We compare our systems to the work of \citet{marcin2018} as they used similar training data. We note, however, that \citet{ge-human} reported even better results with much more (non-public) training data. Comparing \citep{neural-fluency} and \citep{ge-human} suggests that most of their gains come from the larger training set.} (row 1) on CoNLL-2014 and come within 3 GLEU on JFLEG. Since the baseline SMT systems of \citet{marcin2018} were better than the ones we used, we achieve even higher relative gains over the respective SMT baselines (Tab.~\ref{tab:results-smt-compare}).

\begin{table}
\centering
\small
\begin{tabular}{|l|cc|cc|}
\hline
& \multicolumn{2}{c|}{\bf G\&J-D (\citeyear{marcin2018})} & \multicolumn{2}{c|}{\bf This work} \\
 & CoNLL & JFLEG & CoNLL & JFLEG \\
 & (M2) & (GLEU) & (M2) & (GLEU) \\\hline
SMT & 50.27 & 55.79 & 48.18 & 50.86 \\
Hybrid & 56.25 & 61.50 & 58.40 & 58.63 \\\hline
{\bf Rel.\ gain} & {\bf 11.90\%} & {\bf 10.23\%} & {\bf 21.21\%} & {\bf 15.28\%} \\\hline
\end{tabular}
\caption{Improvements over SMT baselines.}\label{tab:results-smt-compare}
\end{table}

\paragraph{Error type analysis}

We also carried out a more detailed error type analysis of the best CoNLL-2014 M2 system with/without training data using ERRANT (Tab. \ref{tab:err_types}). Specifically, this table shows that while the trained system was consistently better than the untrained system, the degree of the improvement differs significantly depending on the error type. In particular, since the untrained system was only designed to handle Replacement word errors, much of the improvement in the trained system comes from the ability to correct Missing and Unnecessary word errors. The trained system nevertheless still improves upon the untrained system in terms of replacement errors by 10 F$_{0.5}$ (45.53 vs.\ 55.63).

In terms of more specific error types, the trained system was also able to capture a wider variety of error types, including content word errors (adjectives, adverbs, nouns and verbs) and other categories such as pronouns and punctuation. Since the untrained system only targets spelling, orthographic and morphological errors however, it is interesting to note that the difference in scores between these categories tends to be smaller than others; e.g. noun number (53.43 vs 64.96), orthography (62.77 vs 74.07), spelling (67.91 vs 75.21) and subject-verb agreement (66.67 vs 68.39). This suggests that an untrained system is already able to capture the majority of these error types.

\begin{table}[t]
  \centering
    \begin{tabular}{|l|r|r|}
\cline{2-3}    \multicolumn{1}{r|}{} & \multicolumn{2}{c|}{\boldmath{}\textbf{ERRANT F$_{0.5}$}\unboldmath{}} \\
    \hline
    \textbf{Type} & \textbf{No train} & \textbf{Train} \\
    \hline
    Missing     & -     & 51.96 \\
    Replacement     & 45.53 & 55.63 \\
    Unnecessary     & -     & 50.38 \\
    \hline
    ADJ   & -     & 27.03 \\
    ADV   & -     & 29.80 \\
    DET   & 19.17 & 55.01 \\
    MORPH & 33.20 & 64.81 \\
    NOUN  & 4.31  & 34.88 \\
    NOUN:NUM & 53.43 & 64.96 \\
    NOUN:POSS & -     & 13.51 \\
    ORTH  & 62.77 & 74.07 \\
    OTHER & 2.45  & 18.39 \\
    PREP  & 34.39 & 56.58 \\
    PRON  & -     & 40.91 \\
    PUNCT & -     & 46.08 \\
    SPELL & 67.91 & 75.21 \\
    VERB  & -     & 37.94 \\
    VERB:FORM & 48.03 & 63.33 \\
    VERB:SVA & 66.67 & 68.39 \\
    VERB:TENSE & 35.39 & 47.90 \\
    \hline
    \end{tabular}%
  \caption{A selection of ERRANT F$_{0.5}$ error type scores comparing the best CoNLL-2014 system with and without training data. A dash means the system did not attempt to correct the error type.}
  \label{tab:err_types}%
\end{table}%

\paragraph{Oracle experiments}

\begin{table}
\centering
\small
\begin{tabular}{|l|c|}
\hline
\textbf{Hypothesis space} & \textbf{Error} \\
 & \textbf{rate} \\
\hline
Expanded input sentence (Tab.~\ref{tab:results-no-smt}) & 61.28\%  \\
SMT lattice (Tab.~\ref{tab:results-smt}, rows 3, 5) & 55.64\%  \\
Expanded SMT lattice (Tab.~\ref{tab:results-smt}, rows 4, 6-8) & 48.17\%  \\
\hline
\end{tabular}
\caption{Oracle error rates for different hypothesis spaces using the first annotator in CoNLL-2014.}\label{tab:oser}
\end{table}

Our FST-based composition cascade is designed to enrich the search space to allow the neural models to find better hypotheses. Tab.~\ref{tab:oser} reports the oracle sentence error rate for different configurations, i.e.\ the fraction of reference sentences in the test set which are not in the FSTs. Expanding the SMT lattice significantly reduces the oracle error rate from 55.63\% to 48.17\%.

\section{Conclusion}

We demonstrated that our FST-based approach to GEC outperforms prior work on LM-based GEC significantly, especially when combined with a neural LM. We also applied our approach to SMT lattices and reported much better relative gains over the SMT baselines than previous work on hybrid systems. Our results suggest that FSTs provide a powerful and effective framework for constraining neural GEC systems.

\section*{Acknowledgements}

This paper reports on research supported by the U.K.\ Engineering and Physical Sciences Research Council (EPSRC grant EP/L027623/1) and Cambridge Assessment, University of Cambridge.

\newpage

\bibliography{naaclhlt2019}
\bibliographystyle{acl_natbib}

\appendix



\end{document}